\documentclass{article}
\usepackage{ijcai23}

\usepackage{times}
\usepackage{soul}
\usepackage{url}
\usepackage[hidelinks]{hyperref}
\usepackage[utf8]{inputenc}
\usepackage[small]{caption}
\usepackage{graphicx}
\usepackage{amsmath}
\usepackage{amsthm}
\usepackage{amssymb}
\usepackage{booktabs}
\usepackage{algorithm}
\usepackage{algpseudocode}
\usepackage[switch]{lineno}

\usepackage{multirow}
\usepackage{subfigure}

\usepackage{lipsum}

\DeclareMathOperator*{\argmax}{arg\,max}

\title{Evolving Dictionary Representation for Few-shot Class-incremental Learning}

\author{
Xuejun Han
\and
Yuhong Guo
\affiliations
Carleton University, Canada
\emails
xuejunhan@cmail.carleton.ca,\;
yuhong.guo@carleton.ca
}

\begin{document}

\maketitle

\begin{abstract}
New objects are continuously emerging in the dynamically changing world and a real-world artificial intelligence system should be capable of continual and effectual adaptation to new emerging classes without forgetting old ones. 
In view of this, 
in this paper we tackle a 
	challenging and practical 
continual learning
	scenario named \emph{few-shot class-incremental learning (FSCIL)}, in which 
labeled data are given for classes in a base session but 
	very limited labeled instances are available for new incremental classes. 
To address this problem, we propose a novel and succinct approach by introducing deep dictionary learning which is a hybrid learning architecture 
that combines
	dictionary learning and visual representation learning
	to provide a better space for characterizing different classes. 
We simultaneously optimize the dictionary and the feature extraction backbone 	
	in the base session,
while only finetune the dictionary in the incremental session 
	for adaptation to novel classes,
	which can alleviate the forgetting on base classes compared to finetuning the entire model.
	To further facilitate future adaptation, we 
also incorporate multiple pseudo classes into the base session training so that certain space projected by dictionary can be reserved for future new concepts.
The extensive experimental results on 
	{\em CIFAR100, {mini}ImageNet} and {\em CUB200} 
	validate the effectiveness 
	of our approach compared to other SOTA methods. 
\end{abstract}

\section{Introduction}

\begin{figure}[t]
    \centering
        \includegraphics[width=0.47\textwidth]{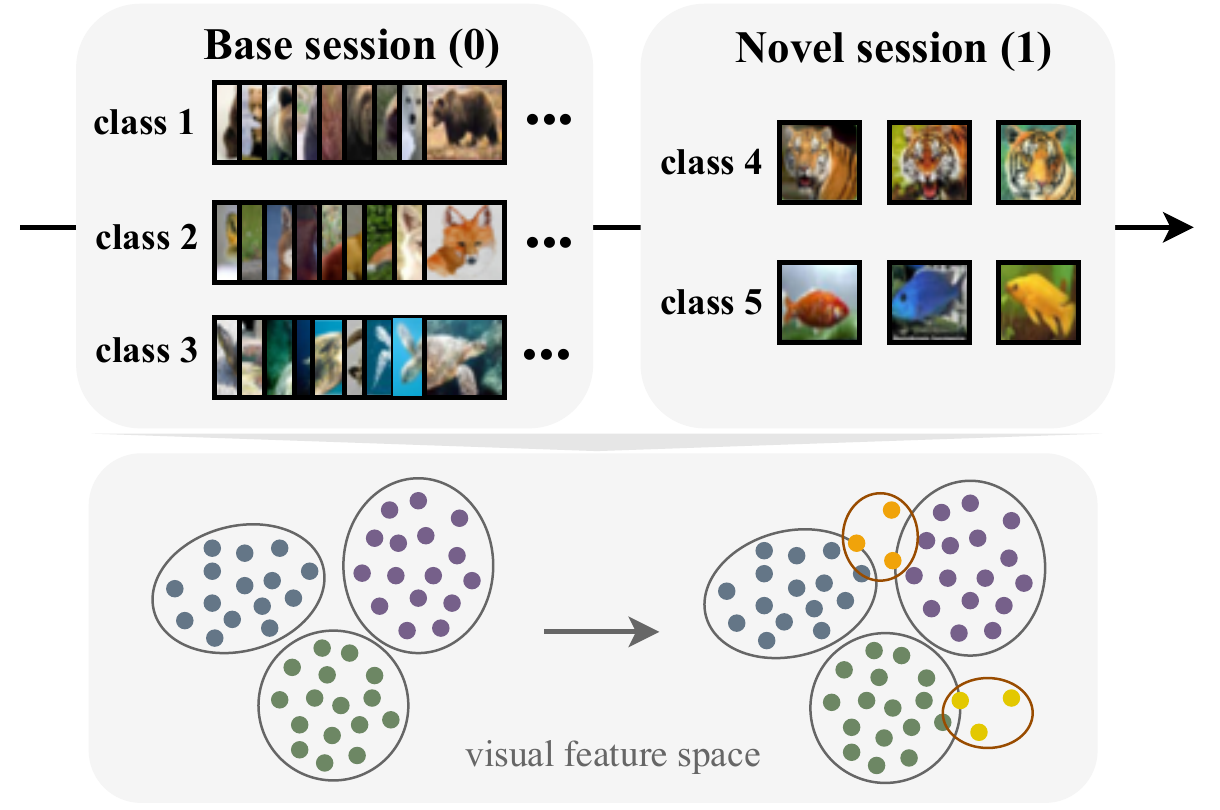}
    \caption{Few-shot class-incremental learning. The model is trained on sufficient labeled data from base classes and extended to cover novel classes with limited labeled data. The goal is to accurately classify all classes via a joint classifier.
    A commonly adopted paradigm to alleviate overfitting to new classes and forgetting on base classes for FSCIL is to freeze the feature extraction backbone after base session and only extend the classifiers's label set. However by merely doing so the class-wise visual  features extracted by backbone for base and novel classes are not well separated, thus leading to performance degradation in novel sessions. }
    \label{fig:fscil}
\end{figure}
Standard machine learning models are normally trained on static datasets and only able to classify the pre-defined categories with identical distributions, which is hardly to adapt to the rapidly changing world where novel objects emerge all the time. Such models are typically fixed upon deployment and therefore lacks the flexibility and compatibility to accommodate new concepts. Moreover, tactless model finetuning on new emerging datasets tends to bring about catastrophic forgetting \cite{CF1,CF2} on what has been learned in the past.
To address this deficiency, the continual or incremental learning \cite{CL,LL} was proposed to enable the model to incrementally learn new information from stream format data. There are three fundamental scenarios in the literature: task-incremental, domain-incremental and class-incremental learning and each of those has its own challenges \cite{IL}. Being the most realistic setting, class-incremental learning (CIL) has been actively explored in resent works \cite{icarl,castro2018,hou2019,wu2019large,pelle2020,scr}, in which the model is trained on a sequence of disjoint sets of classes and a unified classifier is employed for all the experienced classes. Standard CIL setting assumes that there is sufficient labeled data for new classes, which inevitably impedes its practical applications since it is expensive to collect a large amount of annotated data, not to mention that in some cases the unlabeled ones are also hard to obtain, such as images of rare species. Therefore, how to develop an effective incremental model in the absence of sufficient data for novel classes is worth to investigate.

Recently, few-shot class-incremental learning (FSCIL) has been proposed and attracted much interest 
from the research community
\cite{topic,cec,chera2021,zhu2021,peng2022,fact,liu2022} to tackle the problem of data scarcity in CIL. Concretely, the model is initially trained on a large number of annotated data from base classes and then 
incrementally learns new classes with few-shot labeled data. 
There are two challenges involved in FSCIL: the forgetting on base classes due to the inaccessibility to previous data, and the overfitting to few-shot data of new classes as a result of data scarcity in novel sessions. An effective strategy to alleviate the issue of overfitting is to freeze the backbone after finishing the base training \cite{cec}, however this simple scheme 
relies
entirely on base representation learning and lacks the compatibility and extensibility for future adaptation (see Figure \ref{fig:fscil}). \cite{fact} enhances the model compatibility by pre-assigning multiple virtual prototypes to the feature space in the base session but no further session specific adaptation is 
applied to improve the performance over new classes, which results in superior performance on base classes but suboptimal on new ones. 

\begin{figure}[t]
    \centering
        \includegraphics[width=0.45\textwidth]{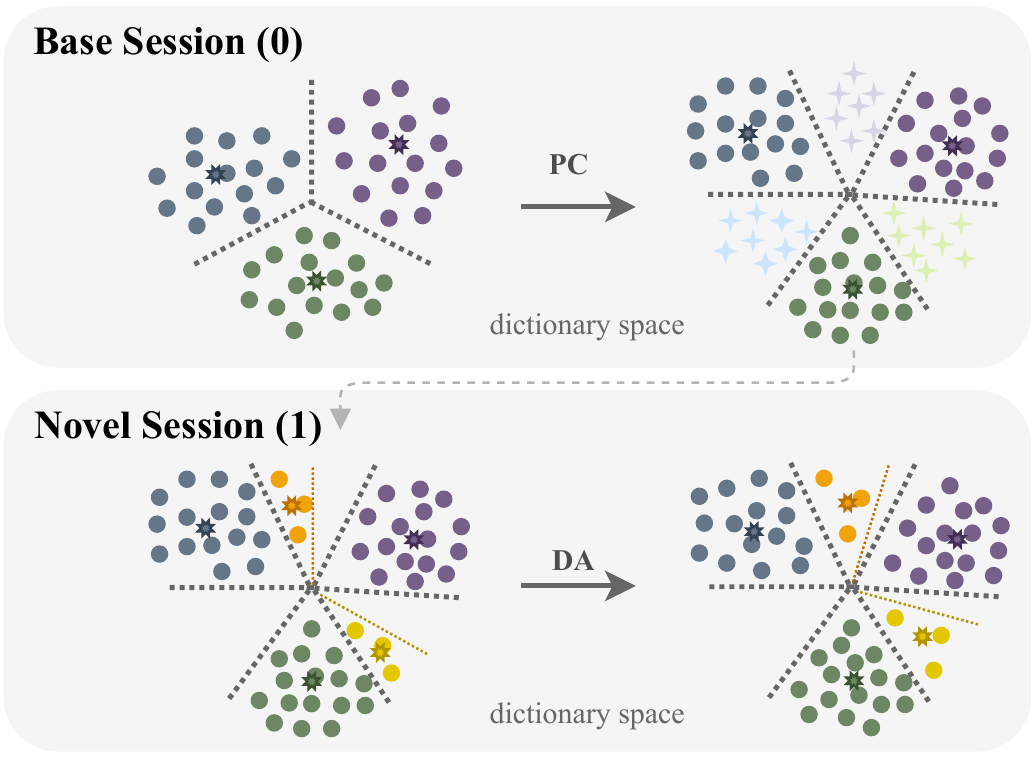}
    \caption{Illustration of our proposed method D-FSCIL. The 4-point stars represent the synthetic data for pseudo classes and the 8-point stars are class prototypes. PC means including pseudo classes in the base session, i.e. pseudo-class augmented learning; DA is the incremental dictionary adaptation to novel sessions. By including pseudo classes into base session training, the class-wise representations possess higher separability, where the intra-class representations become more compact and more free space is reserved for future novel classes. During the phase of dictionary adaptation, base classes' prototypes and backbone are frozen and novel classes' prototypes and dictionary are mildly optimized to better incorporate novel concepts.}
    \label{fig:d-fscil}
\end{figure}

In this paper, we propose a novel and succinct FSCIL method based on deep dictionary learning, which we name by \emph{D-FSCIL} and show in Figure \ref{fig:d-fscil}. 
In this method, the deep feature extractor
and the dictionary representation are trained collaboratively in the base session. 
Meanwhile, we generate 
a set of synthetic data points in the visual feature space as pseudo classes 
to enhance the forward compatibility of the dictionary. By optimizing the backbone and dictionary on extended class set, we expect the data representations of different classes to be more separable and the space reserved by pseudo classes to better incorporate new concepts. The dictionary learned on extended categories in the base session will be more generalizable
and compatible for future adaptation. Afterwards, the backbone is frozen to alleviate the forgetting and overfitting issues and only the dictionary representation is mildly updated on new classes. Extensive experiments on CIFAR100, {mini}ImageNet and CUB200 demonstrate the superiority of our method over the state-of-the-art. 

\section{Related Work}
\paragraph{Few-shot Class-incremental Learning}
Few-shot class-incremental learning as an extension of CIL has attracted a lot of attention recently. \cite{topic} proposes to employ the neural gas structure to preserve the topology of the feature manifold for different classes. \cite{chera2021_cvpr}  introduces a knowledge distillation algorithm for FSCIL and proposes 
to use semantic information during training. \cite{cec} utilizes a graph model to propagate context information between classifiers for adaptation. \cite{liu2022} introduces a data-free replay scheme for generating old samples in FSCIL. \cite{sub} proposes a simple but effective subspace regularizer for learning in 
novel sessions. \cite{metafscil} proposes a bi-level optimization based on meta-learning to directly optimize the network to learn how to incrementally learn.

\paragraph{Dictionary Learning}
Dictionary learning has been widely used in the field of image annotation, multi-label learning, zero-shot learning, transfer learning, etc. Typical dictionary learning is to learn a dictionary/projection matrix to project the visual space to another semantic embedding space \cite{dic2015,zhang2015,ye2017}. In recent years, deep dictionary learning has been actively explored in image classification, image restoration and predictive phenotyping \cite{ddl2019,fu2019,ddl2020,zheng2021}. Different from aforesaid methods, \cite{zhou2021} proposes an end-to-end deep dictionary model 
for multi-label classification
by learning collaborative representations in the projected dictionary space instead of sparse ones. 
In this paper, we adopt a deep dictionary learning architecture to facilitate a better adaptation scheme for FSCIL.

%
\section{Problem Setup}

FSCIL aims to design a model that can incrementally learn novel classes from few-shot data without forgetting knowledge about old classes. Formally, the model experiences a sequence of labeled datasets $\{ \mathcal{D}_0, \mathcal{D}_1,\dots, \mathcal{D}_T\}$ with disjoint classes and can only access to one dataset at a time. After done the training at a time, the data is no longer accessible afterwards. In FSCIL, we call $\mathcal{D}_0=\{(\mathbf{x}_i,y_i )\}_{i=1}^{N_0}$ as the training data for \textbf{base session}, in which a large amount of labeled training data from multiple base
classes is provided. We denote the number of training examples in the base session as $N_0$ and all base classes as $C_0$.
We use $\mathcal{P}_0 =  \{ \mathbf{p}_1^{(0)}, \dots, \mathbf{p}_{C_0}^{(0)} \}$ to denote
all class prototypes to be learned 
for base classes. In the following sessions $\{ \mathcal{D}_1,\dots, \mathcal{D}_T\}$, there is very limited labeled data for each session, which we call \textbf{novel sessions}. In the novel session $t$, we have the few-shot dataset $\mathcal{D}_t=\{(\mathbf{x}_i,y_i)\}_{i=1}^{CK}$ 
in a $C$-way $K$-shot format 
---i.e., $K$ instances from each of the $C$ classes.

\section{Methodology}
The key idea of our method D-FSCIL is to learn a prescient and adaptable deep dictionary learning model to tackle FSCIL. 
We introduce 
a hybrid learning architecture by integrating 
dictionary learning with visual representation learning in Section \ref{sec:ddl}. To further facilitate the future adaptation and improve the generalization and compatibility of the dictionary representation, we generate some pseudo classes by mixing up base data points in the visual feature space, which 
is referred to as 
pseudo-class augmented learning and 
presented in Section \ref{sec:pc}. After done the base session, the backbone is frozen which is proven a simple and effective way to overcome forgetting on past knowledge. Thus in our model, only dictionary is updated to adapt to new classes (see Section \ref{sec:da}). 
Finally, we provide
a detailed description in Section \ref{sec:opt} 
for the model training procedure and inference.

\subsection{Deep Dictionary Learning}\label{sec:ddl}

Intuitively, the idea of deep dictionary learning is to first take advantage of the deep neural network to embed complex input 
data into informative and low-level feature embeddings. After that, the dictionary learning is applied to decompose these embeddings into basic patterns and
re-represent the data in a high-level pattern based space as 
projection coefficients, which 
can be utilized for prediction \cite{fu2019}. In this section, 
we present a deep dictionary learning based classification module for base session training,
in which the deep feature extractor 
and dictionary are trained collaboratively.

We represent the base dataset  $\mathcal{D}_0$ 
as a data matrix $\mathbf{X} =[\mathbf{x}_1,\dots, \mathbf{x}_{N_0}]^\top$ and its 
extracted
visual feature matrix is $\phi(\mathbf{X})\in{\mathbb{R}^{N_0\times d}}$. 
Let $\mathbf{M} \in \mathbb{R}^{m\times d}$ be the dictionary with $m$ atoms 
in the extracted visual feature space, 
and $\mathbf{Z}=[\mathbf{z}_1, \dots, \mathbf{z}_{N_0}]^\top \in \mathbb{R}^{N_0\times m}$ be the coefficient matrix for the $N_0$ instances in the 
base dataset $\mathcal{D}_0$. 
Each coefficient vector $\mathbf{z}_i$ is a high-level dictionary-atom based embedding vector
for $\phi(\mathbf{x}_i)$ in the dictionary space. 
The deep dictionary learning factorizes the visual feature matrix  $\phi(\mathbf{X})$ into the product 
of the dictionary  $\mathbf{M}$ and the coefficient matrix  $\mathbf{Z}$, and learns 
the feature extractor $\phi$, the dictionary $\mathbf{M}$ and the coefficients $\mathbf{Z}$ by minimizing the 
regularized reconstruction error defined as follows,
\begin{equation}\label{loss_dic}
\begin{aligned}
L_{dic} &= \sum_{i=1}^{N_0} \left( \|\phi(\mathbf{x}_i) - \mathbf{z}_i^\top \mathbf{M}\|_2^2  + \lambda \| \mathbf{z_i}\|_2^2 \right)  \\
&
=
\| \phi(\mathbf{X})-\mathbf{Z}\mathbf{M}\|_F^2 + \lambda\| \mathbf{Z}\|_F^2
\end{aligned}
\end{equation}
where $\| \cdot\|_F$ is the Frobenius norm and $\lambda$ is a trade-off parameter.
Different from conventional dictionary learning which learns sparse representations in the dictionary space, 
we adopt the $\ell_2$-norm regularization over $\mathbf{Z}$ to enable
a closed-form solution for $\mathbf{Z}$ given fixed $\mathbf{M}$ and $\phi(\mathbf{X})$
as follows via the minimization of $L_{dic}$,
\begin{equation}\label{get_z}
\mathbf{Z} = \phi(\mathbf{X})\mathbf{M}^\top(\mathbf{M}\mathbf{M}^{\top} +\lambda \mathbf{I}_m )^{-1}
\end{equation}
where $\mathbf{I}_m$ denotes the identity matrix of size $m$.
Henceforward we can treat $\mathbf{Z}$ as a function of 
the dictionary $\mathbf{M}$ and the extractor $\phi$.

With the coefficients $\mathbf{Z}$ for the base session, 
we perform prototype based classification by enforcing the 
coefficient vectors to be close to their corresponding class prototypes 
and minimizing the following
prototype classification loss:
\begin{equation}
L_{cls} = \frac{1}{N_0}\sum_{i=1}^{N_0}  
	\log \frac{\exp(s(\mathbf{z}_i,\mathbf{p}^{(0)}_{y_i})/\tau)}{\sum_{\mathbf{p}\in \mathcal{P}_0} \exp(s(\mathbf{z}_i,\mathbf{p})/\tau)}
\end{equation}
where $s(\cdot,\cdot)$ denotes
the cosine similarity and $\tau$ is a temperature parameter. 
As previously introduced,  $\mathcal{P}_0$ consists of the set of class prototypes in the base session.  

\subsection{Pseudo-Class Augmented Learning} \label{sec:pc}
The dictionary merely learned on the base classes is 
not particularly prescient on 
generalization or adaptation
to new concepts in future novel sessions. Therefore, we propose to
generate some pseudo classes to shed light on
the possible future ones. By this way, the representation space for those pseudo classes can be reserved for 
future incremental classes and the dictionary is expected to be more adaptable and compatible.
Specifically, we extend the base dataset and classes by 
conducting inter-class mixup
over the base training instances
{\em in the visual feature space}
and treat the synthetic data generated by each pair of different classes as a new pseudo class.

Mixup~\cite{Mixup}
is a linear interpolation based data augmentation technique.
We adopt this technique in a particular manner to generate sythetic pseudo-classes. 
Specifically, let $\mathcal{F}_0=\{\mathbf{f}_1,\dots,\mathbf{f}_{N_0} \}$ be the corresponding feature vectors 
for the base training dataset $\mathcal{D}_0$.
To generate the synthetic data for a pseudo-class,
we first randomly sample a pair of base classes, say $c_1$ and $c_2$, and 
denote the pseudo class generated by class $c_1$ and $c_2$ as $\tilde{c}_1$.  
Then, a synthetic feature vector of the pseudo class $\tilde{c}_1$ can be generated by
mixing up two randomly sampled instance vectors in the feature space from class $c_1$ and $c_2$
as follows:
\begin{equation}
\tilde{\mathbf{f}}^{(\tilde{c}_1)} = \gamma \mathbf{f}^{({c}_1)} + (1-\gamma) \mathbf{f}^{({c}_2)}
\end{equation}
where $\mathbf{f}^{({c}_1)}$ and $\mathbf{f}^{({c}_2)}$ are feature vectors from $\mathcal{F}_0$,
belonging to class $c_1$ and $c_2$, respectively. 
We sample the coefficient $\gamma$ from the range of $[0.4,0.6]$ to encourage the synthetic data to fall 
between the two different classes and reduce overlaps with data in existing classes.
By mixing up different pairs of feature vectors from class $c_1$ and $c_2$,
we can generate a set of synthetic feature vectors for the pseudo class $\tilde{c}_1$. 
We repeat such process $\tilde{C}_0$ times by mixing up different pairs of base classes
to get $\tilde{C}_0$ pseudo classes and their corresponding synthetic data. 
Suppose the total number of generated mixup instances 
for the $\tilde{C}_0$ pseudo-classes 
is $\tilde{N}_0$.  
The set of synthetic mixup data in the visual feature space $\phi(\cdot)$ 
can be denoted as $\tilde{\mathcal{F}}=\{ \tilde{\mathbf{f}}_1,\dots, \tilde{\mathbf{f}}_{\tilde{N}_0} \}$
with corresponding labels $\{ \tilde{y}_1,\dots, \tilde{y}_{\tilde{N}_0} \}$,
indicating the pseudo classes.

After obtaining $\tilde{\mathcal{F}}$, its corresponding dictionary based coefficient matrix $\mathbf{\tilde{Z}}$
can be calculated as a function of 
the dictionary $\mathbf{M}$ and the extractor $\phi$
via Eq.~(\ref{get_z}) by replacing $\phi(\mathbf{X})$ with $\tilde{\mathcal{F}}$.
Let $\mathcal{\tilde{P}}_0 =  \{\tilde{ \mathbf{p}}_1^{(0)}, \dots, \tilde{\mathbf{p}}_{\tilde{C}_0}^{(0)} \}$ 
denote 
the set of to-be-learned pseudo class prototypes.
Accordingly, the classification loss on the synthetic data of pseudo classes is defined as:
\begin{equation}
	\tilde{L}_{cls} = \frac{1}{\tilde{N}_0}\sum_{i=1}^{\tilde{N}_0}  
	\log \frac{\exp(s(\tilde{\mathbf{z}}_i,\tilde{\mathbf{p}}^{(0)}_{\tilde{y}_i})/\tau)}{\sum_{\mathbf{p} \in \mathcal{P}_0\cup \mathcal{\tilde{P}}_0} \exp(s(\mathbf{\tilde{z}}_i,\mathbf{p})/\tau) }
\end{equation}
where $\tilde{\mathbf{z}}_i $ is the coefficient vector for $\tilde{\mathbf{f}}_i$ and 
$\tilde{y}_i$ is its corresponding class label among the pseudo classes. 

With the pseudo classes, we then have the following 
total classification loss for the augmented learning of $\mathbf{M}, \phi,$ and class prototypes 
$\mathcal{P}_0\cup \mathcal{\tilde{P}}_0$ 
in the base session:
\begin{equation}\label{loss_l}
	L_{base}= L_{cls} + \eta \tilde{L}_{cls}.
\end{equation}
Through this loss, the dictionary enhanced by pseudo classes is expected to be more generalizable
and capable of effectual adaptation without significantly interfere with the classification of base classes. 

\subsection{Incremental Dictionary Adaptation}\label{sec:da}
In the novel sessions, we freeze the feature extractor $\phi$ 
learned in the base session,
and only finetune the dictionary 
$\mathbf{M}$ to allow convenient 
adaptation to new classes 
without dramatic visual feature shifting. 
Formally, in a novel session $t$, we have a very small labeled dataset $\mathcal{D}_t$ of size $CK$. 
We denote all instances as a data matrix $\mathbf{X}_t$ with instance $\mathbf{x}_{i}$ as 
the $i$-th
row and its visual feature matrix as $\phi(\mathbf{X}_t)\in{\mathbb{R}^{CK\times d}}$. 
The corresponding coefficient matrix is represented by $\mathbf{Z}_t \in\mathbb{R}^{CK\times m}$,
which again can be expressed as a function of $\mathbf{M}$ and $\phi(\mathbf{X}_t)$
via Eq. (\ref{get_z}).
The set of class prototypes to be learned for the current novel session is 
denoted as 
$\mathcal{P}_t =\{ \mathbf{p}^{(t)}_1,\dots,\mathbf{p}^{(t)}_{C}  \}$.

To enable the dictionary to incorporate
information from the current novel session, we slightly update the dictionary $\mathbf{M}$ by minimizing a classification loss on the 
current training dataset, to push the coefficient vectors $\{\mathbf{z}_i\}$ to approach their corresponding prototypes. Specifically, the class prototypes for the 
current session are first initialized as the mean vectors of the coefficients that belonging to the corresponding classes:  
\begin{equation}\label{get_p}
	\mathbf{p}_c^{(t)} = \frac{1}{n_c}\sum_{i=1}^{CK} \mathbb{I}_{[y_i=c]}\mathbf{z}_i
\end{equation}
where $\mathbf{z}_i$ represents coefficient vector in the $i$-th row of $\mathbf{Z}_t$
with corresponding class label $y_i$,
and $n_c$ is the number of instances of class $c$ in $\mathcal{D}_t$. 

Then the dictionary $\mathbf{M}$ and the class prototypes $\mathcal{P}_t$ are learned 
by minimizing the following prototypical classification loss in the novel session $t$:
\begin{equation}\label{l_novel}
\begin{aligned}
	L_{novel}= &\frac{1}{CK}\sum_{i=1}^{CK} \log \frac{\exp(s(\mathbf{z}_i, \mathbf{p}_{y_i}^{(t)} )/\tau) }{\sum_{\mathbf{p} \in \cup_{j \le t} \mathcal{P}_j} \exp (s(\mathbf{z}_i, \mathbf{p})/\tau)}\\
		&+\alpha \| \mathbf{M}-\mathbf{M}_0 \|_F^2
\end{aligned}		
\end{equation}
where $y_i$ denotes the class label for $\mathbf{z}_i$,
$\mathbf{M}_0$ is the dictionary state 
produced from the base session,
and $\alpha$ is a regularization hyper-parameter that controls
the penalty level of dictionary changes. Larger $\alpha$ indicates the higher degree of preserving the dictionary to 
the base session state 
and the lower adaptation to new classes. 
With this regularization term over $M$,
we expect the dictionary representation is only moderately updated to adapt to new classes without severely 
degrading the performance on the base ones.


\begin{algorithm}[tb]
    \caption{Training Procedure}
    \label{alg:algorithm}
    \textbf{Input}: Sequential datasets $\mathcal{D}=\{\mathcal{D}_0,\dots,\mathcal{D}_T \}$ \\
    \textbf{Output}: Learned feature extractor $\phi$, dictionary $\mathbf{M}$,\\ 
	\textbf{\ \ }\qquad\qquad prototypes $\mathcal{P}_0,\dots, \mathcal{P}_T$
        \begin{algorithmic}[1] 
        \State \textbf{	// Base Session}
		\State Randomly initialize $\phi$, $\mathbf{M}$, $\mathcal{P}_0$, ${\mathcal{\tilde{P}}}_0$
        \Repeat
            \State $\mathcal{B} \leftarrow$ sample a mini-batch of instances from $\mathcal{D}_0$;
            \State $\mathcal{B}' \leftarrow$ generate pseudo instances by interclass mixup;
		\State Compute coefficients $\mathbf{Z}$ and $\tilde{\mathbf{Z}}$ for $\mathcal{B}$ and $\mathcal{\tilde{B}}$ by Eq. (\ref{get_z}), respectively;
            \State 
		Update $\phi$, $\mathbf{M}$, $\mathcal{P}_0$, ${\mathcal{\tilde{P}}}_0$ by minimizing $L_{base}$ in Eq. (\ref{loss_l}).
        \Until{maximum iterations reached}\\
        \State \textbf{	// Novel Sessions}
        \State $\mathbf{M}_{0} \leftarrow \mathbf{M}$ and keep $\mathbf{M}_{0}$ fixed
        \For{$t =1,\cdots,T$}
			\State Compute coefficients $\mathbf{Z}_t$ for $\mathcal{D}_t$ by Eq. (\ref{get_z});
                    \State Initialize $\mathbf{P}_t$ 
		    as the means of the coefficient vectors that belonging to the corresponding classes via Eq. (\ref{get_p});
                    \Repeat
                   	 \State Compute coefficients $\mathbf{Z}_t$ for $\mathcal{D}_t$ by Eq. (\ref{get_z});
			\State Update $\mathbf{M}$, $\mathcal{P}_t$ by minimizing $L_{novel}$ in Eq. (\ref{l_novel})
                    \Until{maximum iterations reached}
        \EndFor
    \end{algorithmic}
\end{algorithm}

\subsection{Learning Algorithm and Inference}\label{sec:opt}

\subsubsection{Learning Algorithm}
The proposed deep dictionary learning model D-FSCIL involves the 
minimization of 
the reconstruction loss $L_{dic}$ for dictionary learning implicitly 
during the base or novel session training with the classification loss $L_{base}$ or $L_{novel}$, 
as the coefficient matrices can be expressed as functions of 
the dictionary $\mathbf{M}$ and the extractor $\phi$ via Eq. (\ref{get_z}).
Meanwhile, this also allows the dictionary $\mathbf{M}$ and the extractor $\phi$ 
to be learned through the prototype learning 
classification losses $L_{base}$ and $L_{novel}$ ($\phi$ is fixed for $L_{novel}$).

We describe the detailed model training procedure in Algorithm \ref{alg:algorithm},
which jointly learns $\mathbf{M}$, $\phi$, and  
$\mathcal{P}_0\cup \mathcal{\tilde{P}}_0$ by minimizing $L_{base}$ in the base session,
while jointly learning $\mathcal{P}_t$ and finetuning  $\mathbf{M}$ in each novel session.

\subsubsection{Inference by Prototypes}
Once the learning is finished upon the training set in the session $t$, we obtain the feature extraction module $\phi$, the dictionary $\mathbf{M}$ and prototypes of all seen classes $\cup_{j\le t}\mathcal{P}_j$ for prediction. Given a test 
instance $\mathbf{x}$, its corresponding coefficient vector $\mathbf{z}$ is calculated by Eq. \ref{get_z}, 
and then the label will be assigned to the class with the most similar prototype, 
namely,
\begin{equation}
k^* = \argmax_{ \mathbf{p}_k\in\cup_{j\le t}\mathcal{P}_j } s(\mathbf{z}, \mathbf{p}_k).
\end{equation}

\begin{table*}[t]
    \centering
      \caption{Comparison with the state-of-the-art methods on CIFAR100, \textit{mini}ImageNet and CUB200 datasets. The results of compared methods are 
    directly cited
    from the corresponding papers. The best result in each session and the best average accuracy over all sessions are in bold.}
     \setlength\tabcolsep{4pt}
     \renewcommand{\arraystretch}{0.9}
     
        \resizebox{1.7\columnwidth}{!}{ 
    \begin{tabular}{lccccccccccc} 
        \toprule
        \multirow{1}{*}[-0.8em]{\textbf{Method}} & \multicolumn{9}{c}{ \textbf{Sessions (CIFAR100) w/ ResNet20}} &{\textbf{Average}}    &{\textbf{Final}}   \\
          \cmidrule(lr){2-10}  
         {} & 0 & 1 & 2 & 3 & 4 & 5 & 6 & 7 & 8  &\multirow{1}{*}[0.3em]{\textbf{Acc}} & \multirow{1}{*}[0.3em]{\textbf{Impro.}} \\ 
        \midrule
        TOPIC \cite{topic}   &64.10 &55.88 &47.07 &45.16 &40.11 &36.38 &33.96 &31.55 &29.37  &42.62 &+22.83   \\
         Zhu et al. \cite{zhu2021}	 &64.10 &65.86 & 61.36 &57.34 &53.68 &50.75 &48.58 &45.66 &43.25  &54.51 &+8.95  \\   
       CEC \cite{cec} &73.07 &68.88 &65.26 &61.19 &58.09 &55.57 &53.22 &51.34 &49.14  &59.53  &+3.06   \\
      	  Liu et al. \cite{liu2022} &74.40 &70.20 &66.54 &62.51 &59.71 &56.58 &54.52 &52.39 &50.14 &60.77  &+2.06  \\
	  MetaFSCIL \cite{metafscil} &74.50 &70.10 &66.84 &62.77 &59.48 &56.52 &54.36 &52.56 &49.97 &60.79 &+2.23 \\
	   FACT \cite{fact} &74.60   &72.09   &67.56   &63.52   &61.38   &58.36   &56.28   &54.24   & 52.10  &62.24  &+0.10 \\
	   \cmidrule(lr){1-12}  
	    D-FSCIL (ours)  &\textbf{77.23} &\textbf{73.11} &\textbf{69.11} &\textbf{65.27} &\textbf{62.39} &\textbf{59.48} &\textbf{57.62} &\textbf{55.24} &\textbf{52.20} &\textbf{63.42}\\
        \bottomrule\\
    \end{tabular}
    }
   \medskip

   \resizebox{1.7\columnwidth}{!}{ 
    \begin{tabular}{lccccccccccc} 
        \toprule
        \multirow{1}{*}[-0.8em]{\textbf{Method}} & \multicolumn{9}{c}{ \textbf{Sessions (\textit{mini}ImageNet) w/ ResNet18}} &{\textbf{Average}}    &{\textbf{Final}}    \\
          \cmidrule(lr){2-10} 
         {} & 0 & 1 & 2 & 3 & 4 & 5 & 6 & 7 & 8  &\multirow{1}{*}[0.3em]{\textbf{Acc}} & \multirow{1}{*}[0.3em]{\textbf{Impro.}}\\
        \midrule
        TOPIC \cite{topic}   &61.31 &50.09 &45.17 &41.16 &37.48 &35.52 &32.19 &29.46 &24.42 &39.64    &+26.59  \\
         Zhu et al. \cite{zhu2021}	 &61.45 &63.80 &59.53 &55.53 &52.50 &49.60 &46.69 &43.79 &41.92 &52.75 &+9.09  \\ 
       CEC \cite{cec} &72.00 &66.83 &62.97 &59.43 &56.70 &53.73 &51.19 &49.24 &47.63 &57.75   &+3.38   \\
      	  Liu et al. \cite{liu2022} &71.84 &67.12 &63.21 &59.77 &57.01 &53.95 &51.55 &49.52 &48.21 &58.02  &+2.80  \\
	  MetaFSCIL \cite{metafscil} &72.04 &67.94 &63.77 &60.29 &57.58 &55.16 &52.90 &50.79 &49.19 &58.85  &+1.82  \\
	    FACT \cite{fact} &72.56   &69.63   &66.38   &62.77   &\textbf{60.60}   &\textbf{57.33} &54.34   &52.16   &50.49   &60.69   &+0.52   \\
	   \cmidrule(lr){1-12}  
	    D-FSCIL (ours) &\textbf{74.38} &\textbf{69.76} &\textbf{66.47} &\textbf{62.84} &{59.77} &{56.95} &\textbf{54.50} &\textbf{52.32} &\textbf{51.01} &\textbf{60.83}\\
        \bottomrule\\
    \end{tabular}
    }
      
   \medskip
 \resizebox{1.95\columnwidth}{!}
	{ 
    \begin{tabular}{lccccccccccccc} 
        \toprule
        \multirow{1}{*}[-0.8em]{\textbf{Method}} & \multicolumn{11}{c}{ \textbf{Sessions (CUB200) w/ ResNet18}} &{\textbf{Average}}    &{\textbf{Final}}  \\
          \cmidrule(lr){2-12} 
         {} & 0 & 1 & 2 & 3 & 4 & 5 & 6 & 7 & 8 &9 &10 &\multirow{1}{*}[0.3em]{\textbf{Acc}} & \multirow{1}{*}[0.3em]{\textbf{Impro.}}\\
        \midrule
        TOPIC \cite{topic}   &68.68 &62.49 &54.81 &49.99 &45.25 &41.40 &38.35 &35.36 &32.22 &28.31 &26.28 &43.92 &+31.50    \\
         Zhu et al. \cite{zhu2021}	 &68.68 &61.85 &57.43 &52.68 &50.19 &46.88 &44.65 &43.07 &40.17 &39.63 &37.33 &49.32 &+20.45  \\ 
       CEC \cite{cec} &75.85 &71.94 &68.50 &63.50 &62.43 &58.27 &57.73 &55.81 &54.83 &53.52 &52.28 &61.33 &+5.50      \\
      	  Liu et al. \cite{liu2022} &75.90 &72.14 &68.64 &63.76 &62.58 &59.11 &57.82 &55.89 &54.92 &53.58 &52.39 &61.52  &+5.39  \\
	   MetaFSCIL \cite{metafscil} &75.90 &72.41 &68.78 &64.78 &62.96 &59.99 &58.30 &56.85 &54.78 &53.82 &52.64 &61.92  &+5.14   \\
	   FACT \cite{fact} & 75.90  &73.23   &70.84   &66.13   & 65.56  &62.15   &61.74   & 59.83  &58.41   &57.89   &56.94   & 64.42  &    +0.84\\
	   \cmidrule(lr){1-14}  
	    D-FSCIL (ours)  &\textbf{78.79} &\textbf{75.38} &\textbf{72.05} &\textbf{67.29} &\textbf{66.16} &\textbf{62.42} &\textbf{62.14} &\textbf{60.06} &\textbf{59.25} &\textbf{58.76} &\textbf{57.78} &\textbf{65.46} \\
        \bottomrule
    \end{tabular}
    }
 \label{tab:main}
\end{table*}

\begin{figure*}[ht]
    \centering
    \subfigure[CIFAR100]{
        \includegraphics[width=0.3\textwidth]{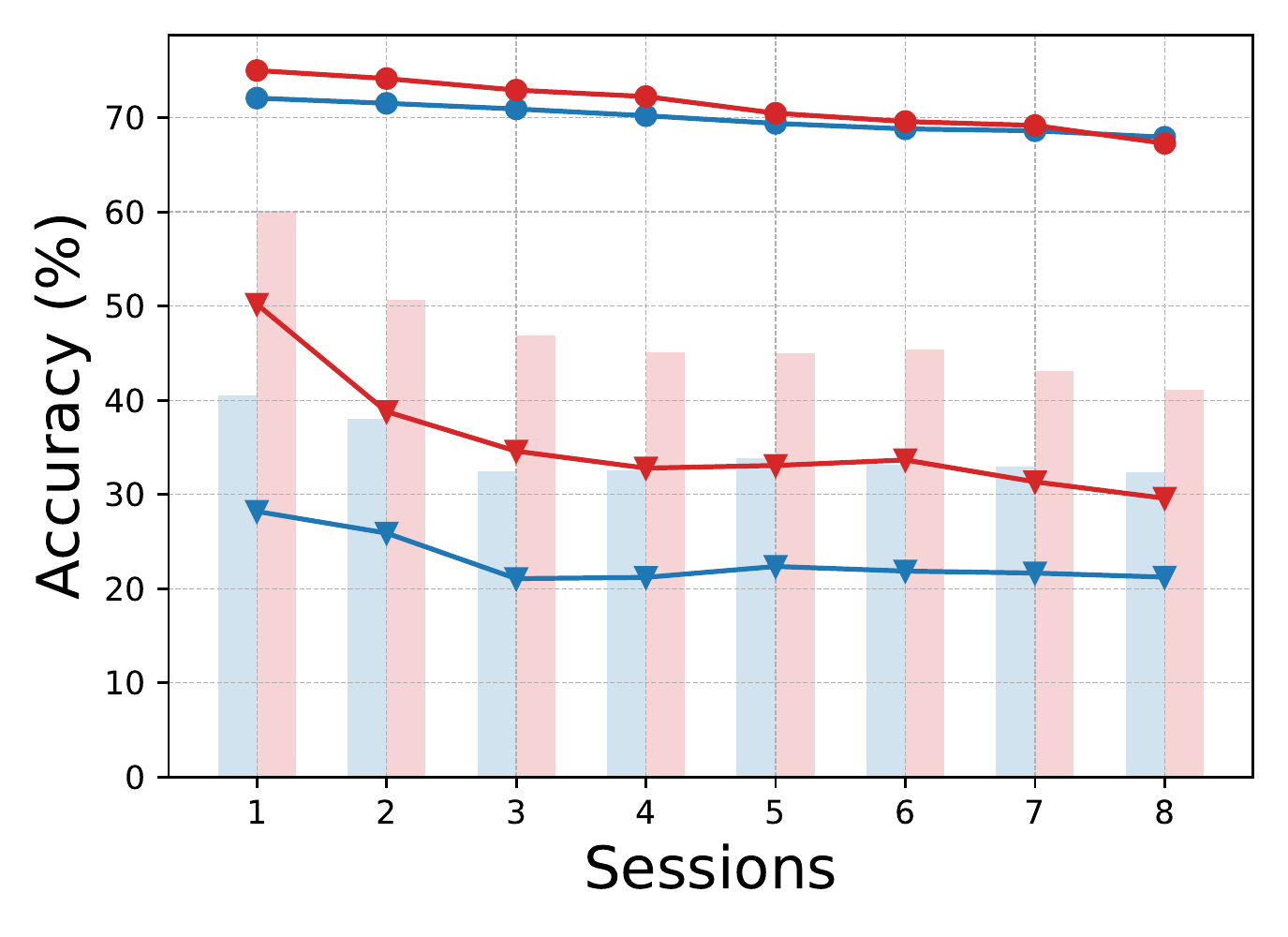}
    }
     \hspace{-0.1in}
    \subfigure[\textit{mini}ImageNet]{
	\includegraphics[width=0.3\textwidth]{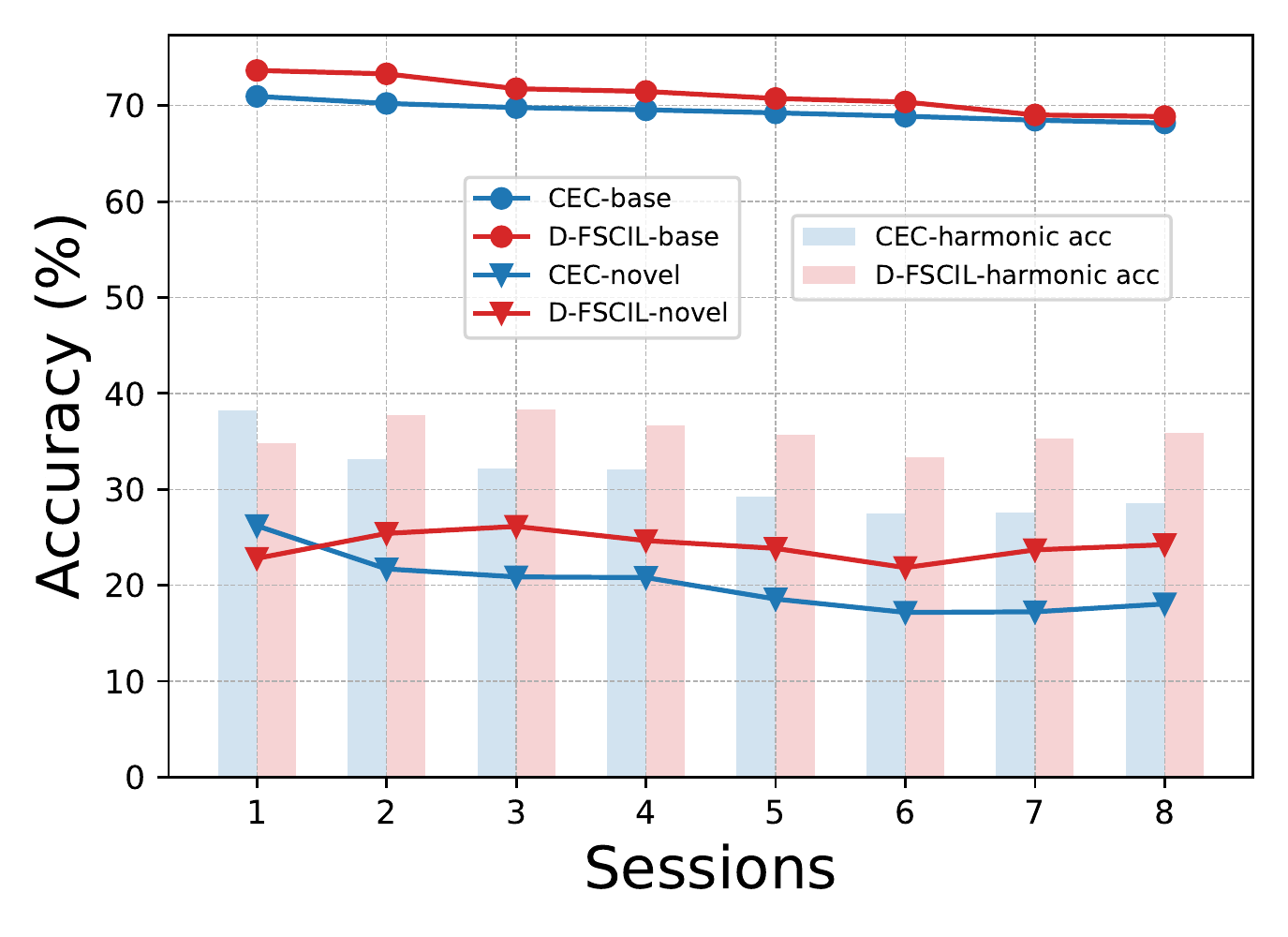}
    }
         \hspace{-0.1in}
    \subfigure[CUB200]{
	\includegraphics[width=0.3\textwidth]{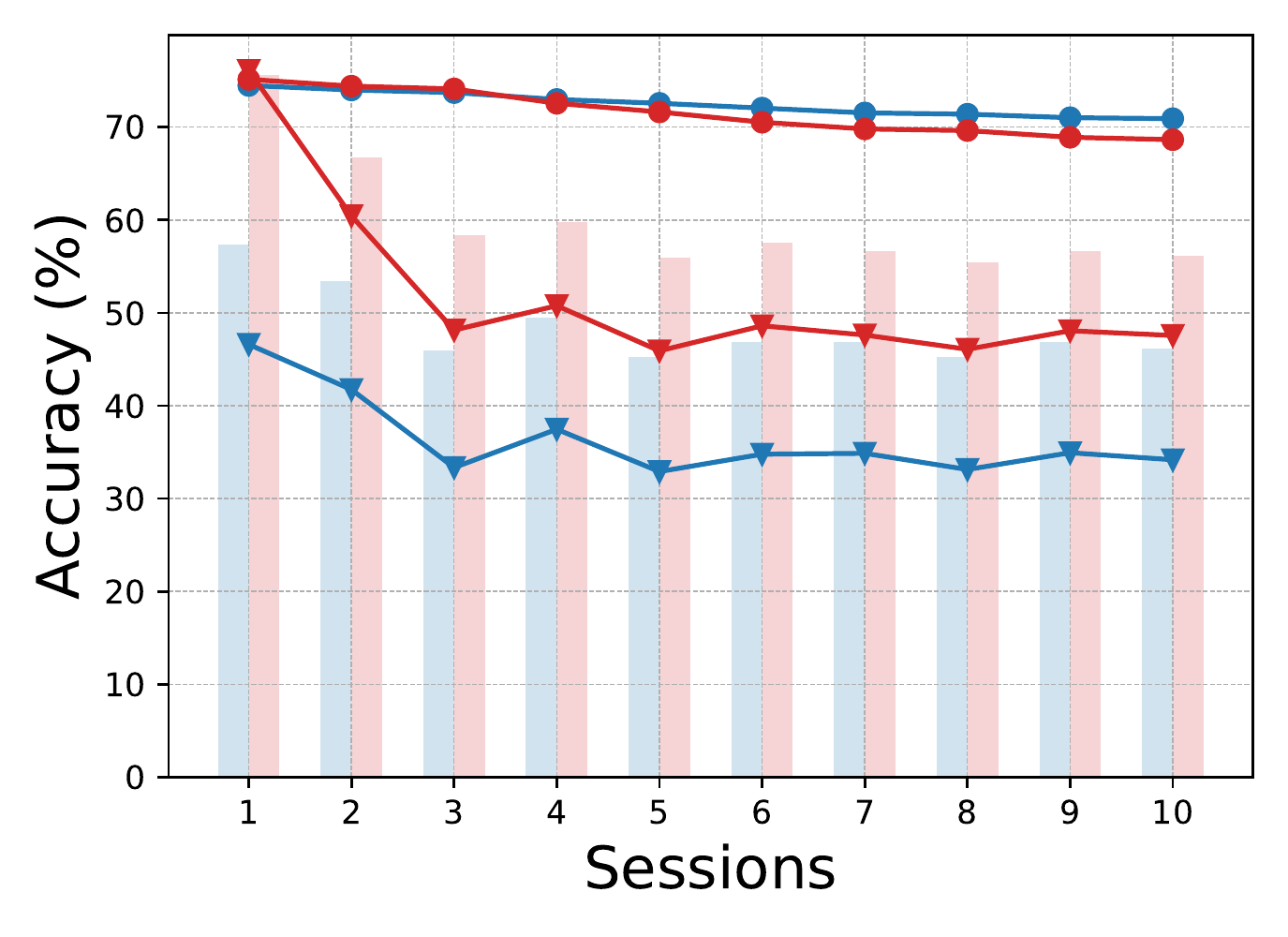}
    }
    \caption{Comparison of CEC (blue) and our method D-FSCIL (red) in terms of the accuracy for base classes and novel classes as well as the harmonic mean in each session on CIFAR100, \textit{mini}ImageNet and CUB200 datasets. Our method D-FSCIL performs particularly well on novel classes while still maintaining satisfactory performance on base classes. }
    \label{fig:harmonic}
\end{figure*}

%
\begin{figure}[ht]
    \centering
        \includegraphics[width=0.3\textwidth]{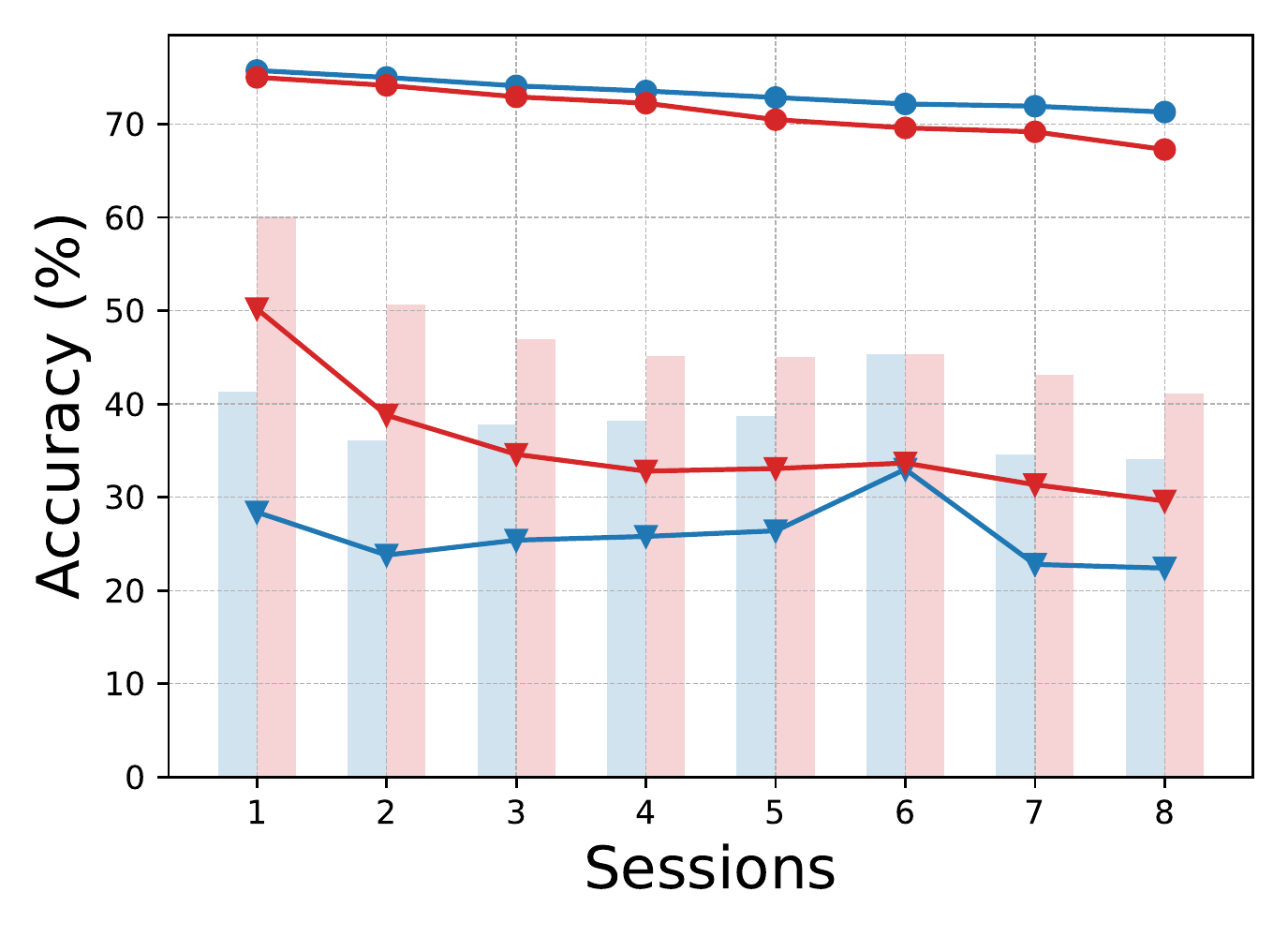}
    \caption{Comparison of FACT (blue) and our method D-FSCIL (red) on CIFAR100. Our method consistently achieves higher harmonic accuracies and better performance on novel classes, while still maintaining satisfactory performance on base classes. }
    \label{fig:fact}
\end{figure}

\begin{figure}[ht]
    \centering
        \includegraphics[width=0.3\textwidth]{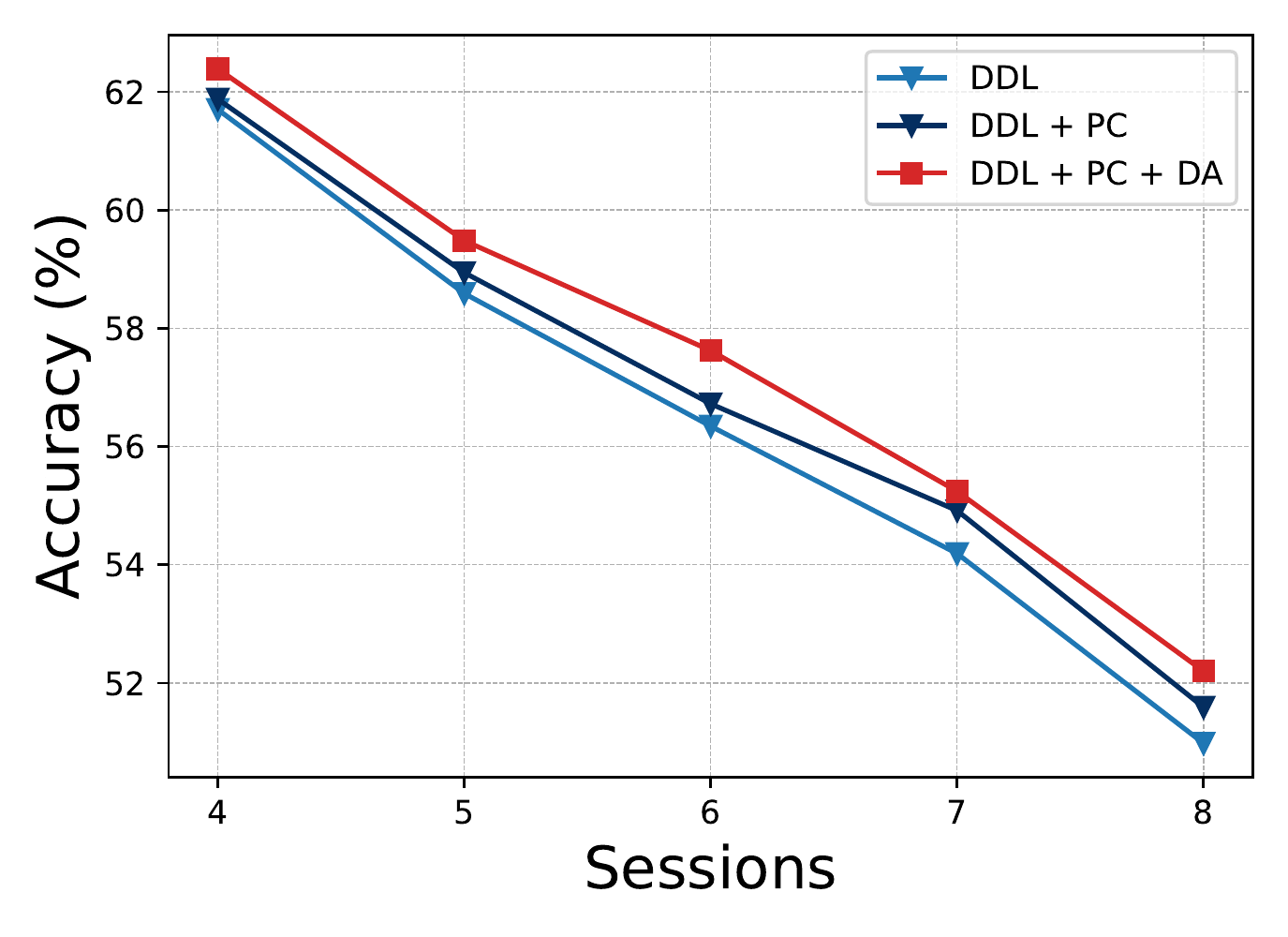}
    \caption{Ablation study on CIFAR100 to verify the effectiveness of each component. DDL represents our base model of deep dictionary learning; PC means including pseudo classes in the base session training; DA is the dictionary adaptation to novel sessions. We show the results in last five sessions for better curves comparison.}
    \label{fig:ablation}
\end{figure}

\begin{figure*}[ht]
    \centering
    \subfigure[influence of $m$]{
        \includegraphics[width=0.25\textwidth]{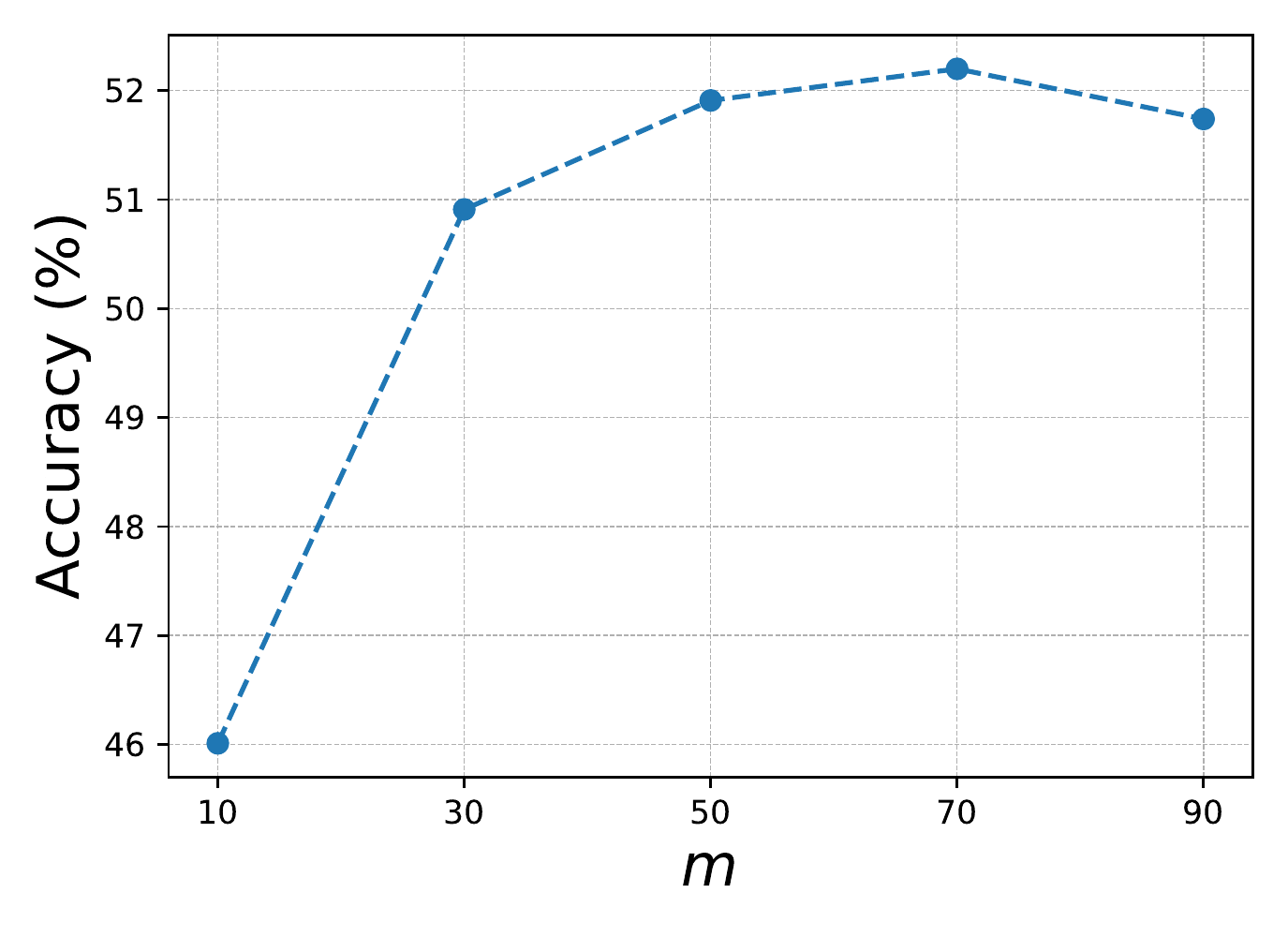}
    }
     \hspace{-0.1in}
    \subfigure[influence of $\lambda$]{
	\includegraphics[width=0.25\textwidth]{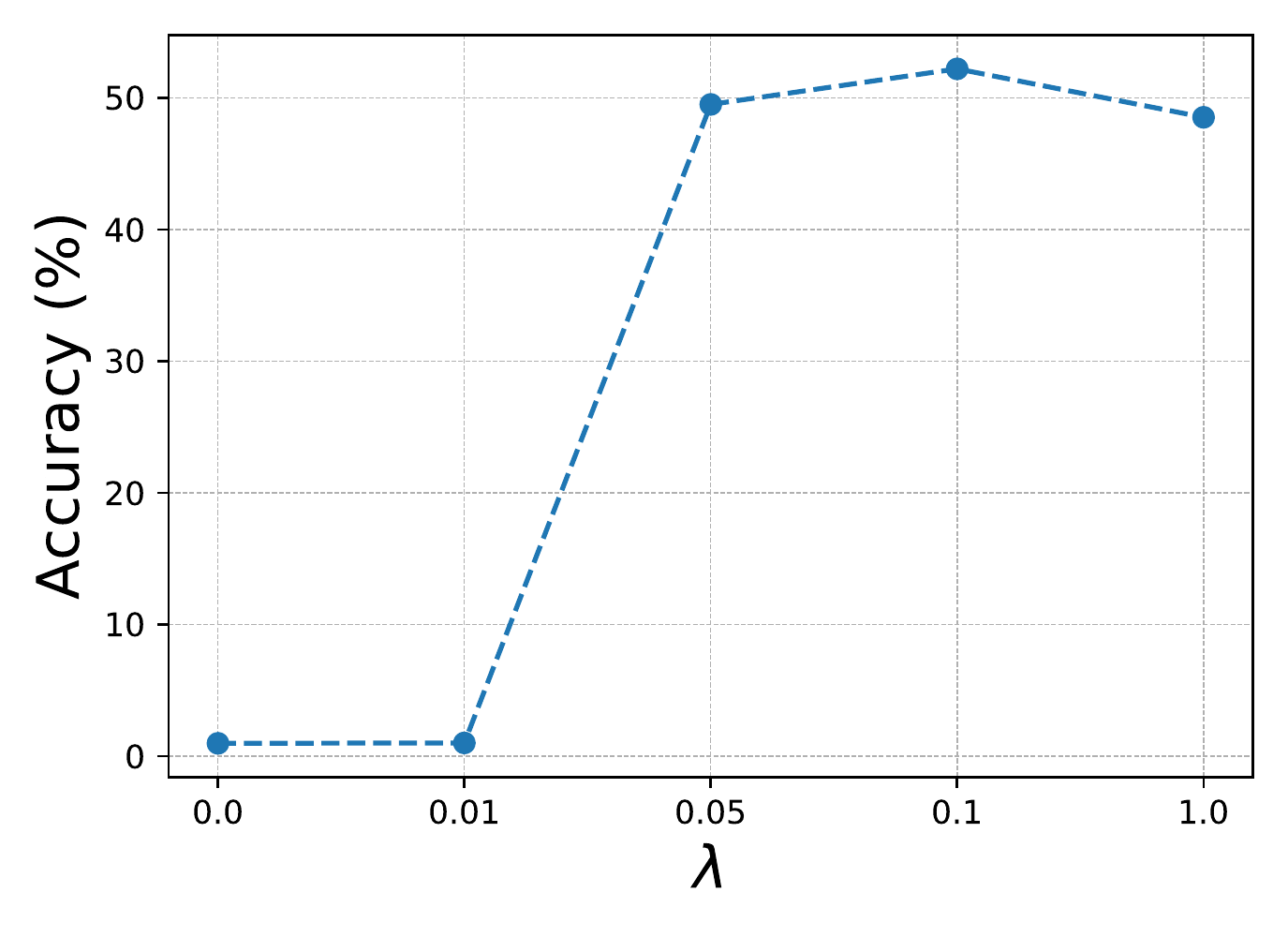}
    }
         \hspace{-0.1in}
    \subfigure[influence of $\tau$]{
	\includegraphics[width=0.25\textwidth]{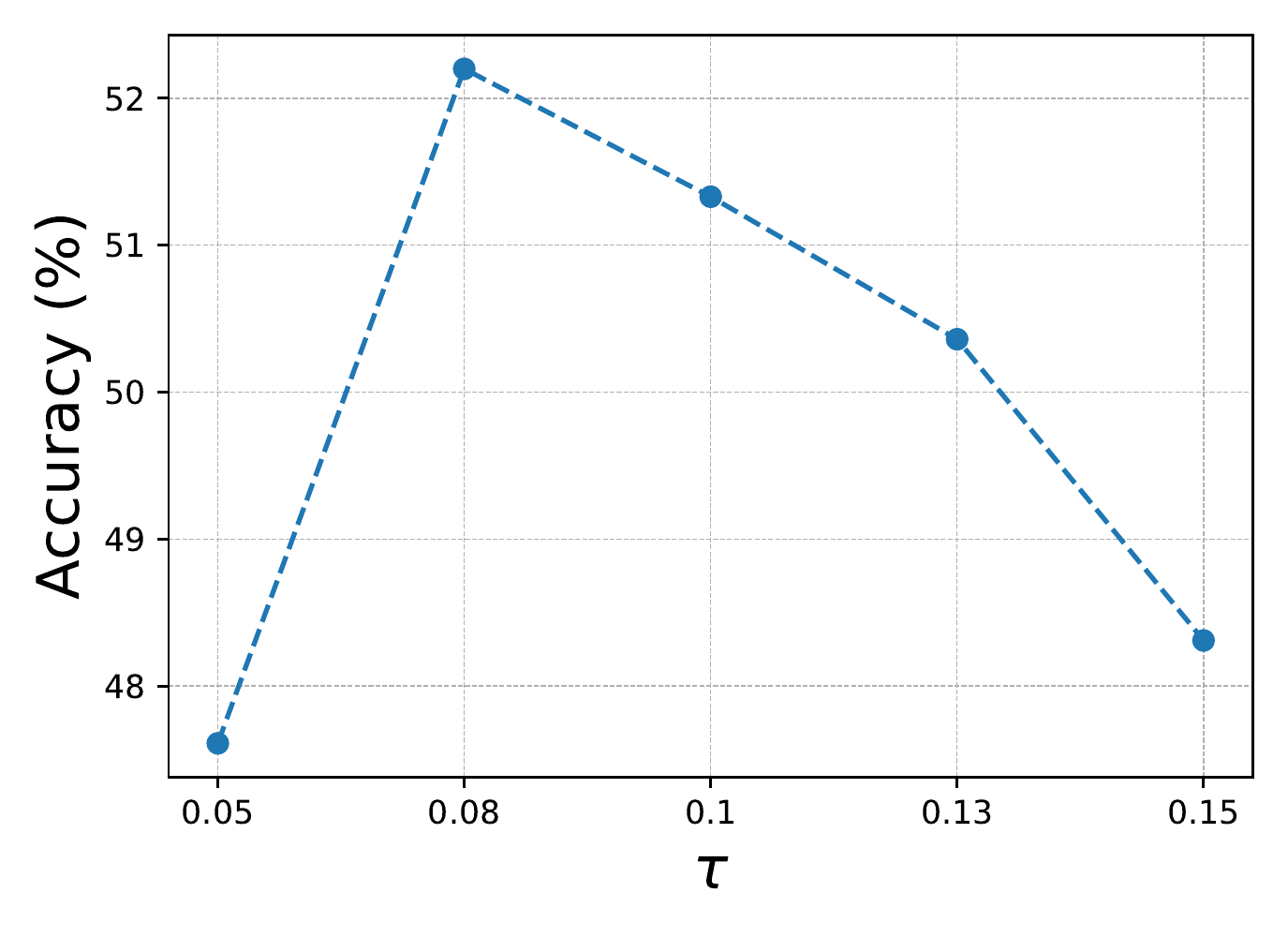}
    }
         \hspace{-0.1in}
    \subfigure[influence of $\eta$]{
    	\includegraphics[width=0.25\textwidth]{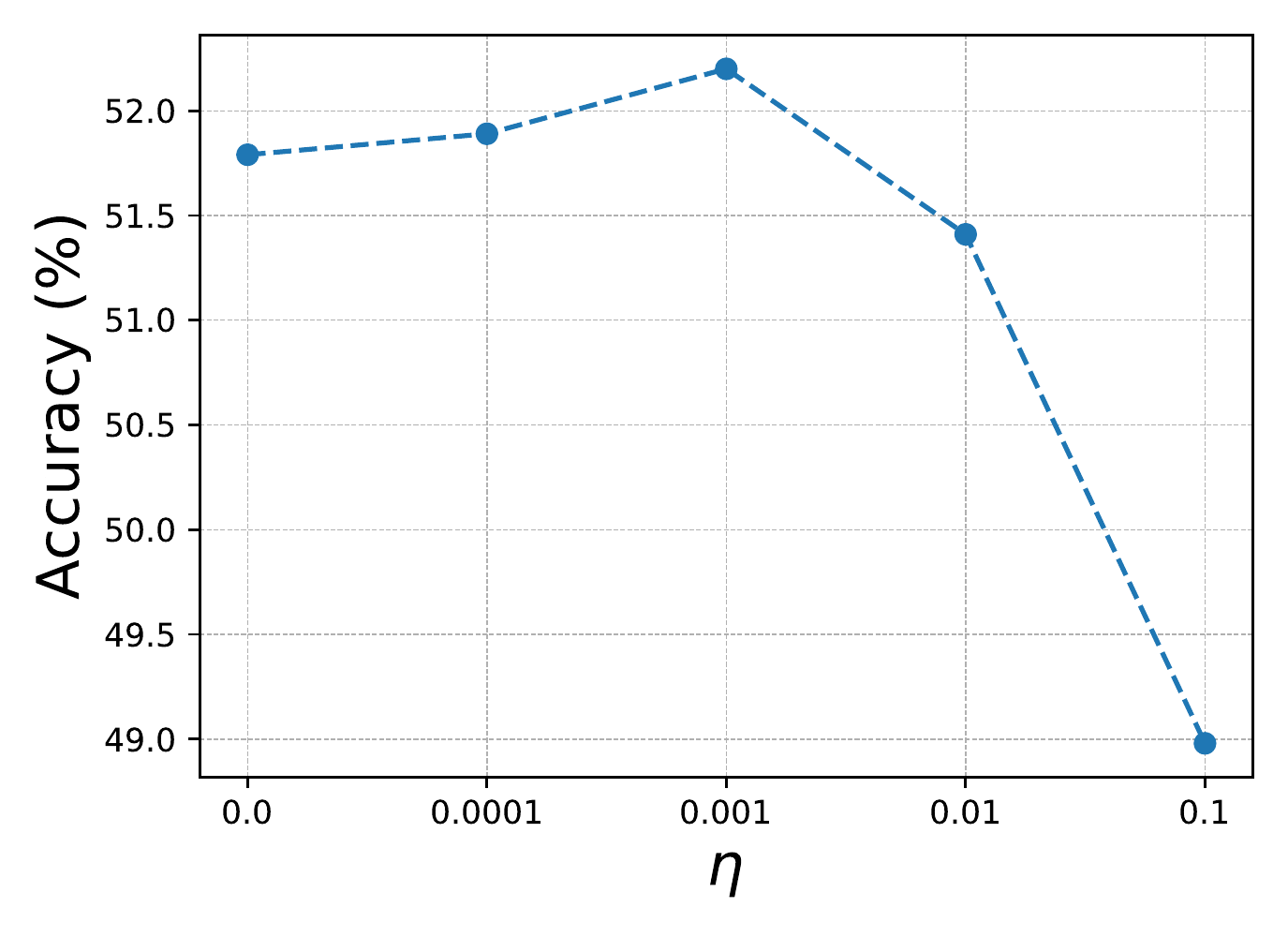}
    }
    \hspace{-0.1in}
    \subfigure[influence of $\alpha$]{
	\includegraphics[width=0.25\textwidth]{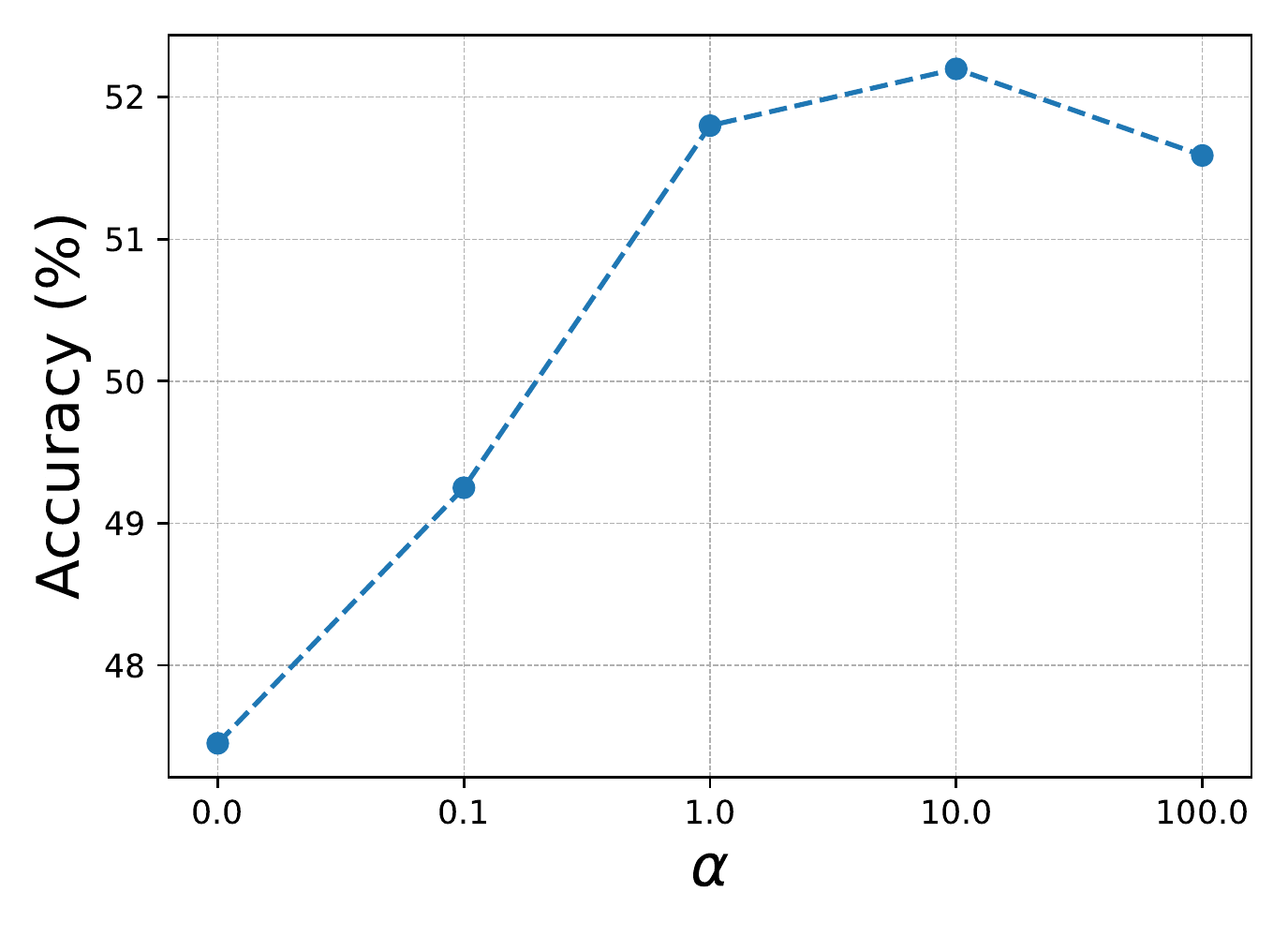}
    }
            \hspace{-0.1in}
    \subfigure[influence of $\tilde{C}_0$]{
	\includegraphics[width=0.25\textwidth]{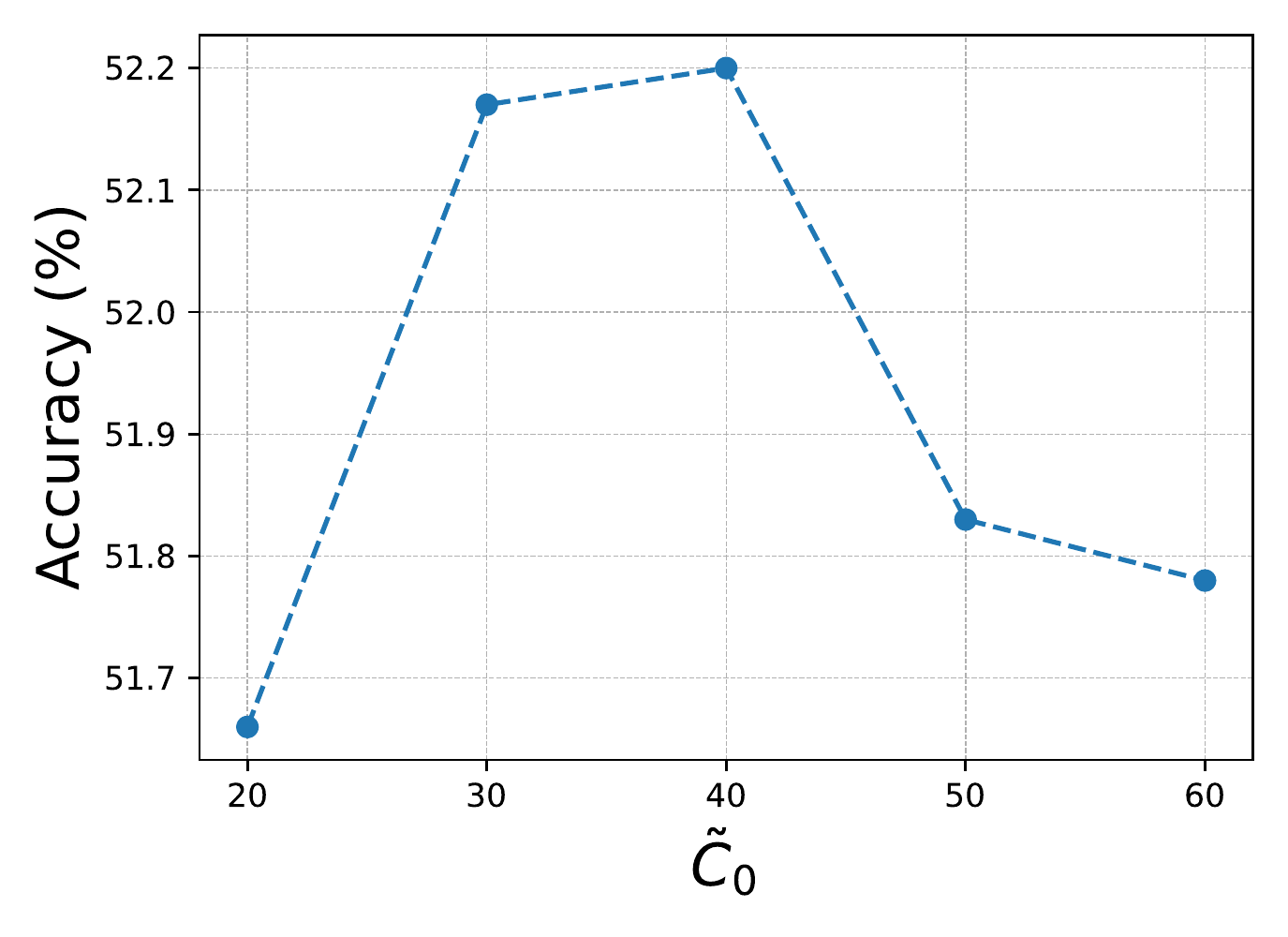}
    }
    \caption{Impact of hyper-parameters on CIFAR100.}
    \label{fig:hyper}
\end{figure*}


\section{Experiments}
We conducted experiments to 
compare the proposed D-FSCIL with several state-of-the-art FSCIL methods on three benchmark datasets. We start by introducing the dataset statistics and experimental implementation details, and then present comprehensive results and analysis on comparison experiments and ablation studies.

\subsection{Datasets}
We follow the benchmark setting in \cite{topic} to evaluate the performance of the proposed method on CIFAR100 \cite{cifar}, \textit{mini}ImageNet \cite{mini} and CUB200 \cite{cub}. 
\textbf{CIFAR100} and \textbf{\textit{mini}ImageNet} consist of 60,000 RGB images from 100 classes. The size of images is 32$\times$32 and 84$\times$84, respectively. Each class contains 500 images for training and 100 images for testing. The total 100 classes are first divided into 60 base classes and 40 novel classes, and the 40 novel classes are further formulated into 8 incremental sessions of \emph{5-way 5-shot} classification task. 
\textbf{CUB200} is a 200-class dataset with 11,788 224$\times$224 RGB images. 100 classes are selected for base session and the rest 100 classes are divided into \emph{10-way 5-shot} tasks for 10 novel sessions. 

\subsection{Implementation Details}
All compared and proposed methods adopt the same backbone architectures. Following \cite{topic}, we employ ResNet20 \cite{resnet} for experiments on CIFAR100 and ResNet18 \cite{resnet} on \textit{mini}ImageNet and CUB200. The model is optimized using SGD with momentum. The batch size is set to be 256 on all datasets. For CIFAR100 and \textit{mini}ImageNet, we train the model for 600 epochs with the learning rate starting from 0.1 and decaying with cosine annealing. On CUB200, the model is trained for 200 epochs and the learning rate starts from 0.005 and decays by a factor of 0.25 every 50 epochs. Data augmentations, such as random crop, random flip and random scale are adopted at base session training time.
In the incremental sessions, the model is finetuned for 10 epochs with learning rate 5e-3 on CIFAR100 and \textit{mini}ImageNet and 1 epoch with learning rate 1e-3 on CUB200.
For the hyper-parameters, $\lambda$ is fixed to be 0.1, $\eta$ is 1e-3 and $\alpha$ is 10 on all datasets.
$\tau$ is set to be 0.08, 0.05, 0.1 for CIFAR100, \textit{mini}ImageNet and CUB200, respectively. 
The size of dictionary $m$ is 70 for CIFAR100 and 500 for \textit{mini}ImageNet and CUB200.
We set the number of pseudo classes $\tilde{C}_0$ as the total number of classes in all the novel sessions. 

\subsection{Evaluation Metrics}
Generally, the top-1 class-wise average accuracy for each session is widely employed for evaluation. As \cite{peng2022} pointed out, such accuracy is not enough for evaluating the overall performance of FSCIL methods since the number of base classes often occupies a large portion of the total. To make up for this deficiency, the \emph{harmonic mean} of accuracies on base classes and novel classes is adopted, which is defined as $2\times A_b\times A_n/ ( A_b+ A_n)$, where $A_b$ and $A_n$ are the accuracy for base classes and novel classes in a session, respectively. An ideal FSCIL method should have high performance in both metrics. Thus, in this paper we will use both to evaluate the proposed and compared methods.

\subsection{Comparison Results}
We compare our proposed D-FSCIL with several state-of-the-art FSCIL methods including \textit{TOPIC} \cite{topic}, \textit{Zhu et al.} \cite{zhu2021}, CEC \cite{cec}, \textit{Liu et al.} \cite{liu2022}, \textit{MetaFSCIL} \cite{metafscil} and FACT \cite{fact}, and the main results are reported in Table \ref{tab:main}. 
In this table, the accuracy for each session is calculated on the test data of all seen classes. We also report the average accuracy over all sessions to reveal the general performance for each method. The last column in the table shows the accuracy increase of our method compared to others after learning all sessions.
It is worth noting that our method D-FSCIL significantly outperforms all the compared state-of-the-art in each session on three datasets. More specifically, D-FSCIL surpasses the most recent methods \textit{Liu et al.}, \textit{MetaFSCIL} and \textit{FACT} in the final session by 2.06\%, 2.23\% and 0.10\% on CIFAR100, 2.80\%, 1.82\% and 0.52\% on \textit{mini}ImageNet, and 5.39\%, 5.14\% and 0.84\% on CUB200, respectively. 

In the base session, our method has higher accuracies on all datasets compared to others demonstrating the superb performance of deep dictionary learning for image classification. Next, we will show that the superior performance of D-FSCIL does not merely benefit from the base classes. On the contrary, our proposed method keeps particular better performance on novel classes after learning all sessions.
Since CEC has comparable results to the most recent methods \textit{Liu et al.} and \textit{MetaFSCIL} with only no more than 1\% accuracy difference, we further compare our method D-FSCIL with CEC in terms of the accuracies on base classes and novel classes, and the harmonic accuracy in each session on all datasets. The results are shown in Figure \ref{fig:harmonic}, from which we can observe that D-FSCIL achieves much better accuracy for novel classes especially on CIFAR100 and CUB200 while still maintaining the competent performance for base classes. The harmonic accuracy of D-FSCIL also consistently outperforms CEC over all sessions for three datasets except the session 1 on \textit{mini}ImageNet where CEC behaves slightly better. 

We further compare our method with FACT \cite{fact} on CIFAR100 since both methods benefit from forward compatibility by embracing pseudo classes in the base session. Notably, FACT forbids the model to further update for novel classes to overcome forgetting on base classes, which indeed achieves such effects but loses better incorporation of new classes. From Figure \ref{fig:fact}, we can see that the good performance of FACT mainly benefits from base classes with less improvements on novel ones, whereas our method can better accommodate novel concepts.

From above observations and analysis, we claim that the proposed method D-FSCIL not only learns the desirable visual feature space and dictionary space for classification in the base session, but also possess the satisfactory capacity to learn and adapt to new classes without forgetting the knowledge of base ones.

\subsection{Ablation Study}

After training the model in the base session, the feature extractor is frozen during learning new sessions and only the dictionary is slightly finetuned to adapt to new classes. In Figure \ref{fig:ablation}, we compare the results of our proposed D-FSCIL (red line), and the results by removing each component including pseudo classes in the base session (PC) and dictionary adaptation to novel sessions (DA). 
We can observe that the deep dictionary learning (DDL) learned by minimizing $L_{dic}$ and $L_{cls}$ can get the decent performance but comparatively lower than other variants. The pure dictionary learning does not take the future possible changes into account and thus the feature extractor and dictionary lack the ability to further generalize to new emerging classes. As a result, it does not reserve any space for new classes in either visual feature space or dictionary space thereby working comparatively poorly. By including the pseudo classes of which the data is generated by inter-class mixup in the base session, the visual representations of the same base class are expected to be closer and the distance between different base classes are far apart, because the pseudo classes are interposed between base classes.
By adding pseudo classes into base session, certain space is reserved for future classes and the dictionary is also trained to map more classes from feature space to dictionary space, therefore, it has higher compatibility to incorporate new classes in novel sessions.
The incremental dictionary adaptation procedure allows the dictionary to moderately change to accommodate novel classes without severely degrading the performance of base classes. In particular, the model learned through pseudo classes has already been growable and provident which can better embrace the dictionary adaptation. 
By combining all the components together, we obtain our model achieving the best results.
Thus far, the ablation studies verified the effectiveness of each component in our proposed D-FSCIL.

\subsubsection{Impact of Hyper-parameters}
There are five hyper-parameters in D-FSCIL, that are the size $m$ of dictionary elements, the trade-off parameter $\lambda$ in the reconstruction loss, the temperature parameter $\tau$ in classification losses, the trade-off parameter $\eta$ to balance the losses over base training data and synthetic data in the pseudo incremental learning, and the parameter $\alpha$ in the incremental dictionary adaptation. 
To evaluate the influence of each hyper-parameter to our method, we conducted experiments on CIFAR100 and the results in the last session with different values of hyper-parameters are shown in Figure \ref{fig:hyper}. It is noted that when evaluating one hyper-parameter, the values of others are fixed.
Firstly, we investigated different values of $m$ from the set of $\{10,30,50,70,90\}$, and we can observe that the performance is relatively poor when $m$ is small like $10$. This is due to the fact that the embedding space projected by the dictionary is excessively compressed because of the low dimensionality, which leads to the loss of data information. The reasonable range of $m$ would be around $70$.
Secondly, the values of $\lambda$ is selected from $\{0, 0.01,0.05,0.1,1 \}$. We can see that no regularization or small values like $0.01$ will fail the deep dictionary learning on CIFAR100 and the best result is produced when $\lambda=0.1$.
Thirdly, the temperature parameter $\tau \in \{0.05,0.08,0.1,0.13,0.15 \}$ and the best result is achieved when $\tau=0.08$ on CIFAR100. 
Fourthly, $\eta \in [0,0.1]$ represents the weight of the classification loss over the generated data of pseudo classes. The larger value of $\eta$ compels the model to focus more on synthetic data leading to the performance deterioration on base classes. $\eta=1e-3$ achieves the best.
Fifthly, $\alpha\in[0,100]$ determines the degree of preserving the dictionary from drastic changes to overcoming forgetting of base classes. $\alpha=0$ yields the worst result because the dictionary is greatly changed to adapt to novel classes. However, due to the face that very limited data is available for novel sessions, the dictionary is easily to overfit the small dataset resulting in the performance degradation in both base and novel sessions.
Lastly, we evaluate the impact of the number of pseudo classes $\tilde{C}_0 \in\{ 20,30,40,50,60\}$ which shows the performance differences are minuscule, thus we default $\tilde{C}_0$ as the total number of novel classes.


\section{Conclusion}
In this paper, we proposed a novel and succinct method D-FSCIL by applying deep dictionary learning into the few-shot class-incremental learning (FSCIL) problem. In the base session, the dictionary and feature extraction module are collaboratively learned
implicitly through the prototypical classification loss minimization. 
For a more adaptable and prescient dictionary, we generate some pseudo classes into base training set by 
conducting inter-class mixup in the visual feature space. 
The representation space occupied by pseudo classes is expected to be reserved for future novel classes
and increase the generalizability. 
Incremental dictionary adaptation in novel sessions further improves the compatibility with new concepts
through dictionary finetuning. Finally extensive experiments have been conducted 
and the results verify the effectiveness of the proposed D-FSCIL.

\bibliographystyle{named}
\bibliography{paperbib}
\end{document}